%% file: root.tex
\documentclass[letterpaper, 10 pt, conference]{ieeeconf}  % Comment this line out if you need a4paper

\IEEEoverridecommandlockouts                              % This command is only needed if 
\usepackage{hyperref}
\usepackage{graphicx} 
\usepackage{multirow}
\usepackage{caption}
\usepackage{algorithm}      
\usepackage{algpseudocode}
\usepackage{amsmath,amssymb} 
\usepackage[table,xcdraw]{xcolor}
\usepackage{stfloats}
\usepackage{booktabs}
\usepackage{multirow}
\usepackage{pifont}
\usepackage{threeparttable}
\usepackage{booktabs}
\usepackage{fontawesome5}
\usepackage{xcolor}

\newcommand{\highlightg}[1]{%
  \begingroup
  \setlength{\fboxsep}{0pt}%
  \colorbox{gray!20}{\strut #1}%
  \endgroup
}

\title{\LARGE \bf
History-Conditioned Spatio-Temporal Visual \\ Token Pruning for Efficient Vision-Language Navigation
}

\author{Qitong Wang$^{1}$, Yijun Liang$^{2}$, Ming Li$^{2}$, Tianyi Zhou$^{3}$ and Christopher Rasmussen$^{1}$
\thanks{Corresponding author: cer@cis.udel.edu}% <-this % stops a space
\thanks{Project Page: https://wqtwjt1996.github.io/publications/2026-vln.html}%
\thanks{$^{1}$The authors are with the Department of Computer and Information Sciences at the University of Delaware, Newark, DE 19711, USA.
{\tt\small wqtwjt@udel.edu; cer@cis.udel.edu}}%
\thanks{$^{2}$The authors are with the University of Maryland's Department of Computer Science, College Park, MD 20742, USA.
{\tt\small yliang17@umd.edu; minglii@umd.edu}}%
\thanks{$^{3}$The author is with the Mohamed bin Zayed University of Artificial Intelligence, Masdar City, Abu Dhabi 7909, UAE.
{\tt\small Tianyi.Zhou@mbzuai.ac.ae}}%
}
\begin{document}

\maketitle
\thispagestyle{empty}
\pagestyle{empty}

%%%%%%%%%%%%%%%%%%%%%%%%%%%%%%%%%%%%%%%%%%%%%%%%%%%%%%%%%%%%%%%%%%%%%%%%%%%%%%%%

\input{tex/0_abstract}
\input{tex/1_introduction}
\input{tex/2_related_work}
\input{tex/3_methodology}
\input{tex/4_experiments}
\input{tex/5_conclusion}

%%%%%%%%%%%%%%%%%%%%%%%%%%%%%%%%%%%%%%%%%%%%%%%%%%%%%%%%%%%%%%%%%%%%%%%%%%%%%%%%

\bibliographystyle{IEEEtran}
\bibliography{IEEEabrv, IEEEtranBST/IEEEfull}

\end{document}

%% file: tex/0_abstract.tex
\begin{abstract}

Vision-Language Navigation (VLN) enables robots to follow natural-language instructions in visually grounded environments, serving as a key capability for embodied robotic systems. 
Recent Vision-Language-Action (VLA) models have demonstrated strong navigation performance, but their high computational cost introduces latency that limits real-time deployment. 
We propose a training-free spatio-temporal vision token pruning framework tailored to VLA-based VLN. 
We apply spatial token selection to the current view, alongside spatio-temporal compression for historical memories, enabling efficient long-horizon inference while reducing redundant computation. 
Leveraging attention-based token importance and query-guided spatio-temporal filtering, the proposed approach preserves navigation-relevant information without retraining or modifying pretrained models, allowing plug-and-play integration into existing VLA systems. 
% Experiments on standard VLN benchmarks show that our method yields superior navigation accuracy—especially under extreme pruning ratios, while maintaining highly competitive real-time inference efficiency compared to existing strategies.
Through experiments on standard VLN benchmarks, we confirm that our method significantly outperforms existing pruning strategies. 
It successfully preserves superior navigation accuracy under extreme pruning scenarios, all while maintaining the highly competitive inference efficiency.
Real-world deployment on a Unitree Go2 quadruped robot further validates reliable and low-latency instruction-following navigation under practical robotic constraints.
We hope this work helps bridge the gap between large-scale multimodal modeling and efficient, real-time embodied deployment in robotic navigation systems.

\end{abstract}

%% file: tex/1_introduction.tex
\section{Introduction}

Vision-Language Navigation (VLN) is a central problem in robotics, aiming to let embodied agents follow natural-language instructions through visually perceived environments. 
It supports practical applications where non-expert users can command robots for tasks such as household assistance, office guidance, and search-and-rescue missions~\cite{vln1}. 
In parallel, recent Vision-Language-Action (VLA) models~\cite{rt2, openvla, pi0} have shown strong effectiveness for robot control, translating vision-and-language understanding into actions for embodied tasks, and are increasingly being adopted for navigation settings, including legged locomotion and instruction-following navigation. 
However, VLA models are typically transformer-based and computationally heavy, and VLN often requires real-time, closed-loop decision making on robots; this creates a latency–responsiveness tension that can hinder reliable deployment of VLAs in VLN~\cite{liu2024revisiting}.

% \begin{figure}[t]
%     \centering
%     \includegraphics[width=\columnwidth]{fig_tab/fig_1_vis.pdf}
%     \caption{
%     % \Ming{Is it possible to add a visulization of the selected tokens? Like the fig 1 of SparseVLM, and the fig 1 of VisPruner? Otherwise, we can also draw a plot translate form Table 1. The x axis is the pruning ratio, and the y axis is the performance, to show our method is more stable and better than the existing methods.}
%     % }
%     % \Qitong{Working on it now. :)
%     \textbf{Illustration of different vision token pruning methods for Vision–Language Action (VLA) models for navigation.}
%     At a pruning ratio of XX\%, our method (d) exhibits more effective token selection than other approaches (a–c), focusing on navigation-critical objects such as XXX. 
%     Importantly, these objects are explicitly mentioned in the instruction prompt, demonstrating that our method better preserves task-relevant visual information.
%     }
%     \label{fig1}
% \end{figure}

\begin{figure}[t]
    \centering
    \includegraphics[width=\columnwidth]{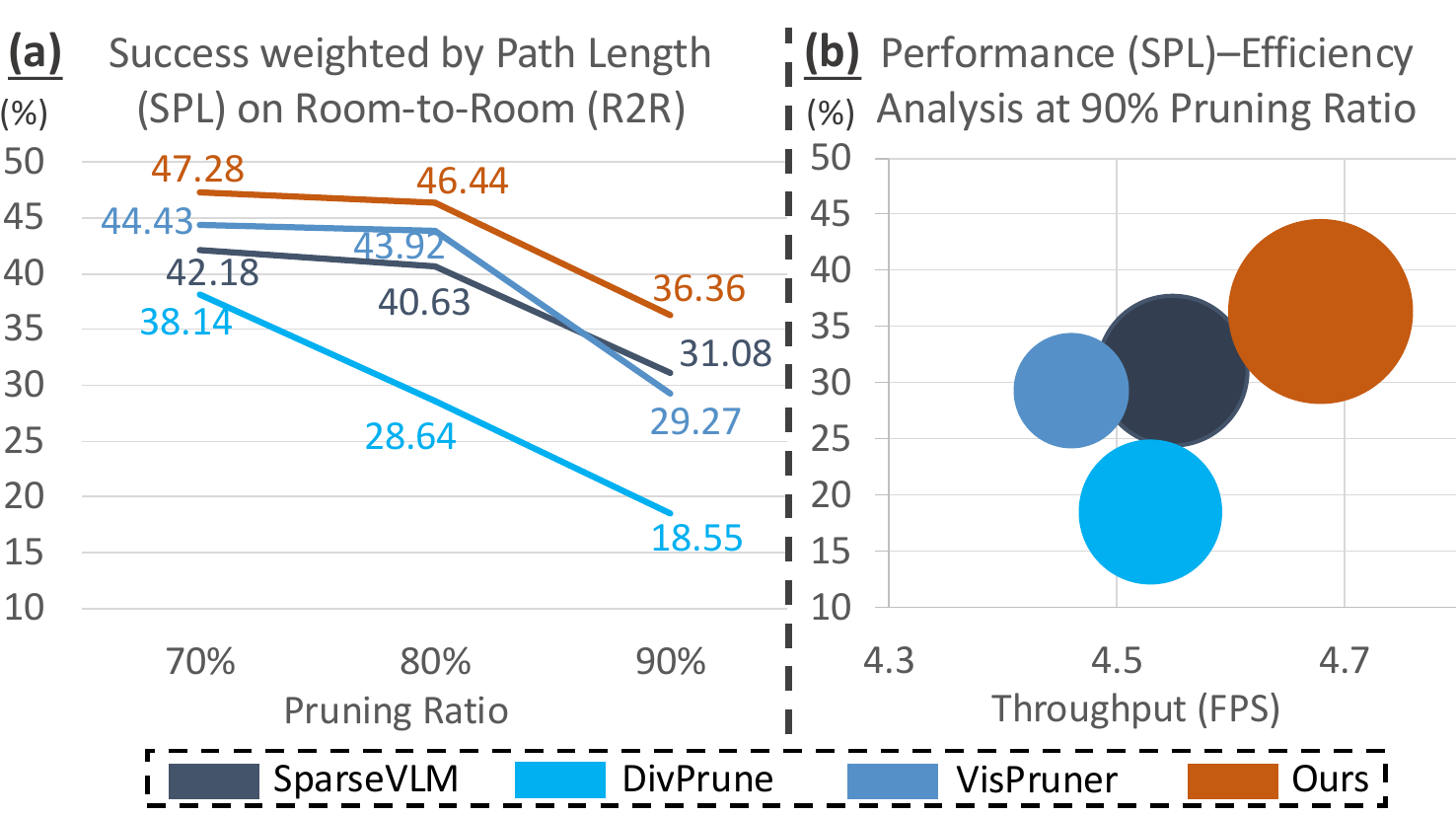}
    \caption{
    \textbf{Performance stability and efficiency–performance dynamics of different vision token pruning methods.}
    (a) SPL performance on the R2R benchmark across pruning ratios from 70\% to 90\%. 
    While all methods experience performance degradation under more aggressive pruning, our method consistently maintains higher SPL than others (SparseVLM, DivPrune, and VisPruner), demonstrating superior preservation of critical visual information.
    (b) Performance–efficiency analysis at 90\% pruning ratio. 
    The horizontal axis shows inference throughput (FPS), and the vertical axis shows SPL performance. 
    Bubble size represents the reduction in CUDA inference latency (ms) compared to the unpruned model, where larger bubbles indicate greater efficiency gains. 
    % Our method achieves both the highest throughput and strongest performance while also providing the largest latency reduction, establishing a superior performance–efficiency trade-off.
    % Our method achieves both the highest throughput and strongest performance while also providing the largest latency reduction, delivering a superior performance–efficiency profile.
    Our method achieves both the highest throughput and strongest performance while maintaining competitive latency reduction, ultimately delivering a superior performance-efficiency profile.
    }
    \label{fig1}
\end{figure}

A promising direction to mitigate this latency bottleneck is \textit{vision token pruning}~\cite{sparsevlm, divprune, vispruner}, which accelerates inference by reducing the number of visual tokens processed while maintaining accuracy as much as possible. 
Yet, token pruning for \textit{VLA models specifically under the VLN setting} remains relatively limited compared with general-purpose vision transformer acceleration. 
Meanwhile, VLN has a task-specific structure: it is commonly modeled as a partially observable decision process where the agent leverages historical observations~\cite{navila, streamvln} (not just the current frame), and effective policies must reason over the spatio-temporal relationship between past and current views. 
These characteristics make ``pruning for VLN'' meaningfully different from pruning in single-image or purely reactive settings. 
Therefore, our core research question is: \textit{How can we perform vision token pruning for VLA models effectively in vision-language navigation, while preserving the spatio-temporal information needed for history-conditioned VLN reasoning?}

In this work, we introduce a \textit{training-free spatio-temporal visual token pruning} approach tailored to VLN in recent VLA-style navigation frameworks where ``historical observations are critical'' for long-horizon decision making.
Our key idea is to treat ``current'' and ``history'' frames differently: we preserve \textit{spatial} coverage for the current observation, while \textit{spatio-temporally} compressing historical memories to reduce redundant computation.
Concretely, we propose an \textit{Adaptive Maximal Marginal Relevance (A-MMR)} strategy, an advanced variant of traditional MMR~\cite{mmr}, to dynamically extract a highly representative subset of visual tokens.
Unlike standard approaches that rely on hard-coded token splits, A-MMR utilizes a unified iterative formulation to dynamically balance saliency and diversity. 
This ensures that while prioritizing semantically rich, high-attention foreground objects, the iteratively selected tokens remain semantically diverse and non-redundant, effectively capturing a comprehensive representation of salient features.
Furthermore, we extend this efficiency to history frames via a \textit{Query-Guided Re-weighting} mechanism. 
We reweight historical tokens based on their relevance to the pruned current-frame queries.
Applying the same A-MMR selection to these re-weighted tokens ensures we construct a highly compact, yet comprehensively informative memory pool.
Finally, by avoiding pruning-time retraining, our proposed token pruning method follows the spirit of plug-and-play efficiency methods that aim to cut latency without shifting pretrained representations, which is especially desirable for embodied transfer.

We conduct extensive experiments on well-established VLN benchmarks~\cite{r2r, rxr}. 
The results show that our approach outperforms existing training-free vision token pruning methods~\cite{sparsevlm, divprune, vispruner} by a substantial margin. 
% Moreover, our method achieves superior efficiency compared to prior pruning strategies~\cite{sparsevlm, divprune, vispruner}, establishing a favorable trade-off between efficiency and effectiveness.
Moreover, our method achieves superior efficiency compared to prior pruning strategies~\cite{sparsevlm, divprune, vispruner}, largely mitigating the traditional compromise between efficiency and effectiveness.
For example, on the Room-across-Room (RxR)~\cite{rxr} with 90\% pruning ratio, our method outperforms existing methods, including SparseVLM~\cite{sparsevlm}, DivPrune~\cite{divprune}, and VisPruner~\cite{vispruner}, by 12.04\%, 18.35\%, and 7.57\%, respectively, when evaluated using Success weighted by Path Length (SPL). 
In terms of efficiency, under 90\% pruning ratio, our method reduces the CUDA inference latency from 231.34\, ms (unpruned) to 213.40\, ms, outperforming SparseVLM, DivPrune, and VisPruner by 6.09\, ms, 7.31\, ms, and 10.96\, ms.
Lastly, we demonstrate real-world deployment on a Unitree Go2 quadruped robot, verifying the practical applicability of our method in embodied navigation scenarios.
% \Ming{Need to add more details about the improvements on performance and efficiency, both on benchmarks and real-world deployment. We need to let the readers know how our method outperforms the existing methods after reading the introduction.}
% \Qitong{Fixed both here and later. Note that there is no quantitative numbers for real-world deployment, which contains visualizations only (image or video). This configuration follows the codebase (StreamVLN) we used.}

To summarize, our contribution is three-fold:

$\bullet$ We study an under-explored problem: how to perform efficient vision token pruning for VLA-based vision-language navigation while preserving the spatio-temporal information required for history-conditioned decision making.

$\bullet$ We propose a training-free spatio-temporal token pruning framework that explicitly distinguishes spatial token selection for the current frame and spatio-temporal memory compression for history frames, enabling efficient long-horizon navigation without any training.

% $\bullet$ Experiments on standard VLN benchmarks validate that our method achieves superior performance and efficiency compared with existing vision token pruning methods, and real-world deployment on a Unitree Go2 quadruped robot further demonstrates its practical applicability. \Ming{Same as above, need specific values to let the readers know how much we improve.}

$\bullet$ Experiments on standard VLN benchmarks validate that our method achieves superior performance and efficiency compared with existing vision token pruning methods (e.g., up to 17.81\% SPL and 10.96\, ms latency improvement at 90\% pruning on Room-to-Room (R2R) dataset~\cite{r2r}), and real-world deployment on a Unitree Go2 quadruped robot further demonstrates its practical applicability.

%% file: tex/2_related_work.tex
\section{Related Work}

\subsection{Vision-Language-Action Models}

Vision-Language-Action (VLA) models treat robot control as a unified multimodal sequence modeling problem, mapping visual observations and language instructions to executable actions for embodied applications such as manipulation, mobile navigation, and long-horizon household tasks. 
Recent progress started from RT-2~\cite{rt2}, which showed that web-scale vision-language knowledge can be transferred to robotic action generation by tokenizing actions in a VLM-style framework. 
Building on this paradigm, OpenVLA~\cite{openvla} provided a large open-source 7B VLA trained on 970k real-world robot episodes, substantially improving accessibility and adaptation for downstream robot setups. 
In parallel, policy architectures have expanded beyond autoregressive decoding to generative action modeling: 
$\pi_0$~\cite{pi0} introduced a flow-matching VLA design for general robot control, and subsequent variants (e.g., $\pi_{0.5}$~\cite{pi0_5}) further emphasized long-horizon reasoning and heterogeneous data utilization. 
Recent work also explores native unified token spaces across vision, language, and action (e.g., UniVLA~\cite{univla}) and systematic fine-tuning strategies for VLA adaptation, indicating a shift from ``single-policy training'' toward scalable pretrain-then-adapt ecosystems. 
Overall, current VLA research is moving toward more general, open, and transferable robot foundation policies, while key challenges remain in data efficiency, embodiment transfer, and robust deployment in dynamic real-world environments.

\subsection{Vision-Language Navigation}

Building on the broader VLA paradigm, Vision-Language Navigation (VLN) studies how an embodied agent grounds natural-language instructions into sequential navigation actions. 
Early progress was driven, in part, by benchmark construction, especially Room-to-Room (R2R)~\cite{r2r}, which established photo-realistic instruction-following navigation in Matterport3D~\cite{matterport3d}, and Room-Across-Room (RxR)~\cite{rxr}, which scaled VLN to multilingual, longer-horizon, and densely grounded instruction trajectories.
With stronger foundation backbones, recent VLN research has shifted from task-specific encoders toward unified vision-language(-action) policies that better support long-horizon inference and deployment efficiency. 
In this trend, NaVILA~\cite{navila} formulates legged-robot VLN as a hierarchical VLA pipeline, where high-level language-conditioned motion intents are generated first and then executed by locomotion control, bridging instruction understanding and robot embodiment.
More recently, StreamVLN~\cite{streamvln} introduces a fast dialogue context plus memory context to improve low-latency action generation under long-horizon interaction.
These models indicate a clear direction for modern VLN: moving towards VLA-based, temporally aware, and deployment-oriented embodied navigation systems.

\begin{figure*}[t]
\vspace*{0.05in}
    \centering
    \includegraphics[width=\textwidth]{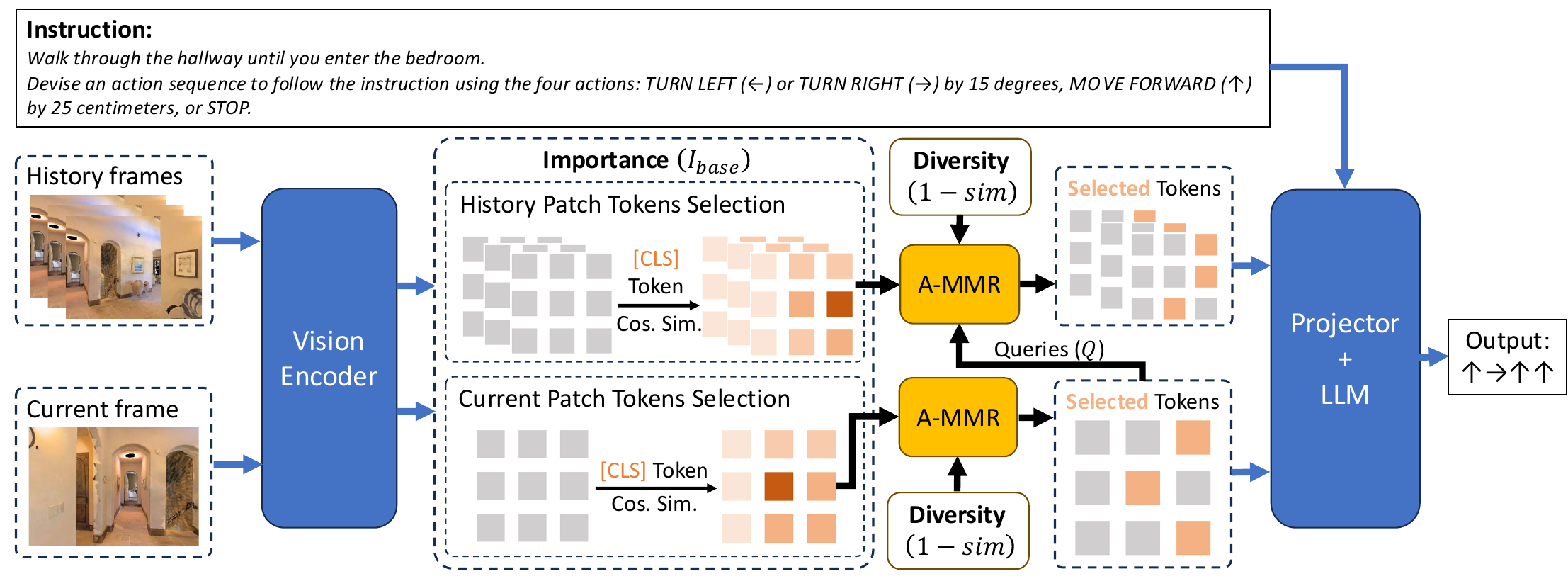}
    \caption{
    % \textbf{Overview of our proposed token pruning framework.} 
    % The vision-language action model takes a natural language instruction and visual observations (comprising history frames and current frames) as input. 
    % Our core methodology relies on the \textbf{Adaptive Maximal Marginal Relevance (A-MMR)} strategy:
    % \textbf{(1) Feature Extraction \& Importance Calculation:} 
    % Visual observations are first processed by a Vision Encoder. We compute the base importance ($I_{base}$)~\ref{formula:I_base} of the visual patches by extracting the attention weights from the global \texttt{[CLS]} token. 
    % \textbf{(2) Token Selection (Current):} 
    % In the bottom branch, an iterative algorithm selects tokens by balancing their base importance and spatial diversity ($1-sim$). 
    % These Selected Tokens serve as Queries ($Q$) for the historical context.
    % \textbf{(3) Token Selection (History):} 
    % In the top branch, history tokens are iteratively selected based on their importance, diversity ($1-sim$), and temporal relevance guided by the Queries ($Q$)  from the current frame.
    % \textbf{(4) Action Prediction:} 
    % Finally, the temporally compressed and spatially preserved selected tokens are processed by a projector and Large Language Model (LLM) to output the final navigation action sequence (e.g., $\uparrow \rightarrow \uparrow \uparrow$).
    \textbf{Overview of our proposed token pruning framework.}
    Given a natural-language instruction and visual observations (history frames and the current frame), our framework proceeds in four stages.
    \textbf{(A) Feature extraction and importance computation:}
    All frames are encoded by a vision encoder, and the base importance $I_{\text{base}}$ of each patch token is computed from the attention weights of the global \texttt{[CLS]} token (normalized with Eq.~\ref{formula:I_base}).
    \textbf{(B) Token selection (current):}
    We apply \textit{Adaptive Maximal Marginal Relevance (A-MMR)} to select current-frame tokens by jointly considering base importance (i.e. semantics) and spatial diversity $(1-\textit{sim})$.
    The selected current tokens are used as queries $Q$.
    \textbf{(C) Token selection (history):}
    We again apply \textit{A-MMR} to select history tokens based on importance and diversity, additionally guided by spatio-temporal relevance conditioned on the queries $Q$ from the current frame.
    \textbf{(D) Action prediction:}
    The selected tokens are fed into a projector and an LLM to predict the final navigation action sequence (e.g., $\uparrow \rightarrow \uparrow \uparrow$).
    % \Ming{In the caption, A-MMR is written as including the (1234) steps, but in the figure, A-MMR is independent of the steps. It's strange.}
    % \Qitong{Fixed.}
    }
    \label{fig_method}
\end{figure*}

\subsection{Vision Token Pruning}
Deploying VLA policies on real robots requires low-latency inference under limited onboard compute and resource budgets. 
Since visual tokens dominate the input length and transformer cost, vision token pruning becomes a practical strategy to reduce computational complexity and inference latency while preserving navigation performance.
SparseVLM~\cite{sparsevlm} introduces a training-free, text-guided pruning paradigm that scores visual tokens via cross-modal (image-text) attention and progressively removes low-importance tokens, with token recycling to mitigate information loss. 
While effective, this relevance-driven selection can retain many mutually similar tokens, yielding redundancy. 
At high compression ratios, such redundancy may prevent keeping enough \emph{unique} tokens to faithfully represent the original visual content.
Motivated by this limitation, DivPrune~\cite{divprune} formulates token pruning as a ``max-min diversity optimization'' problem, explicitly maximizing diversity among retained tokens to reduce redundancy and improve coverage of the original token set, especially under aggressive pruning.
% More recently, VisPruner~\cite{vispruner} combines semantic importance and diversity-aware deduplication in a plug-and-play manner: it both selects informative tokens based on visual attention cues and removes near-duplicate tokens through similarity filtering.
More recently, VisPruner~\cite{vispruner} relies on a predefined selection heuristic. 
It preserves the representative visual subset for model inference by splitting the focus between tokens that are visually informative and those that are semantically distinct.
% This design can be viewed as integrating the semantic selection spirit of SparseVLM with the diversity objective emphasized by DivPrune.
Despite strong progress in general vision token pruning, dedicated studies for VLN remain limited. 
Existing methods are largely frame-centric and do not explicitly model VLN-specific spatio-temporal structure. 
For example, VLN decision making benefits from historical observations~\cite{navila, streamvln} (not only the current frame).
Therefore, how to prune history-frame tokens while preserving long-horizon instruction grounding is still underexplored. 
% Spatially, current pruning criteria are usually token-level saliency/similarity measures and do not explicitly preserve navigation-critical geometric cues (e.g., free space, traversability boundaries, doorway/turn landmarks, and object-instruction spatial relations), which are crucial for accurate action selection in embodied navigation.

%% file: tex/3_methodology.tex
\section{Methodology}

Although recent systems~\cite{navila, streamvln} show that historical observations are crucial for robust long-horizon decision making, existing methods remain largely frame-centric and fail to exploit VLN-specific spatio-temporal redundancy. 
Motivated by this gap, we develop a spatio-temporal pruning formulation tailored to VLN. 
Our approach explicitly considers both current-frame (spatial) and history-frame (spatio-temporal) token utility under long-horizon instruction grounding to significantly improve inference efficiency. 

In this paper, we study \textit{training-free vision token pruning} to avoid modifying pretrained model parameters. 
This prevents distribution shifts caused by training-time pruning or fine-tuning, which can alter the model's internal features and hurt generalization in embodied navigation settings.

The overall framework of our method is illustrated in Figure~\ref{fig_method}. 
It consists of the following four steps:

% \subsection{Definition of Token Importance (Stage 1 in Figure~\ref{fig_method})}
\subsection{Feature Extraction and Importance Computation}
Both history and current frames are first processed by the VLA model’s vision encoder to produce visual tokens, which are subsequently selected by our method. 
Concretely, we first estimate patch importance to rank tokens before pruning. 
Instead of averaging attention scores across all heads, following~\cite{vispruner}, we determine patch importance using the output representations from the vision encoder. 
Specifically, we compute the cosine similarity between the global \texttt{[CLS]} token and the spatial patch tokens to serve as our effective attention weights. 
Since the \texttt{[CLS]} token serves as the aggregate representation of the image, this similarity distribution naturally highlights semantically salient regions (e.g., navigation goals, obstacles).

Let $\mathbf{A} \in \mathbb{R}^{N}$ denote the attention map of the \texttt{[CLS]} token with respect to $N$ patch tokens. 
We normalize these weights to $[0, 1]$ to obtain the base importance score $I_{base}$:
\begin{equation}
    I_{base}^{(i)} = \frac{\mathbf{A}^{(i)} - \min(\mathbf{A})}{\max(\mathbf{A}) - \min(\mathbf{A}) + \epsilon},
\label{formula:I_base}
\end{equation}
where $\mathbf{A}^{(i)}$ represents the attention weight of the $i$-th token, and $\epsilon$ is a small constant added for numerical stability to avoid division by zero (set to $10^{-6}$ in our implementation).

% Our method is readily applicable to VLAs with different vision encoders. 
For VLAs whose vision encoders lack an explicit class token, the global semantic representation can be approximated by aggregating the encoder outputs, such as through mean pooling or self-attention-based aggregation, following existing practices~\cite{mpvg, hiprune}.

% \subsection{Phase I: Current Frame Pruning (Stage 2 in Figure~\ref{fig_method})}
\subsection{Token Selection (Current)}
% For the current observation, we aim to preserve a comprehensive spatial representation. 
% Building upon the traditional MMR~\cite{mmr}, we propose an \textit{Adaptive Maximal Marginal Relevance (A-MMR)} strategy. 
% By evaluating tokens via a unified formulation, $Score = Attention \times (1 - Similarity)$, A-MMR bypasses the need for hard-coded token splits. 
% It dynamically balances the selection process, naturally transitioning from sampling high-attention objects to capturing diverse background representations.
% The selection process is iterative: at each step, we select a token $i$ from the candidate set $\mathcal{U}$ that maximizes the following objective:
% \begin{equation}
%     i^* = \operatorname*{argmax}_{i \in \mathcal{U}} \left( I_{base}^{(i)} \cdot \left( 1 - \max_{j \in \mathcal{S}} \operatorname{sim}(\mathbf{f}_i, \mathbf{f}_j) \right) \right),
% \label{eq:mmr_score}
% \end{equation}
% where $\mathcal{S}$ is the set of already selected tokens, $\mathbf{f}$ denotes the feature vector, and $\operatorname{sim}(\cdot, \cdot)$ is the cosine similarity.
To preserve a comprehensive spatial representation for the current observation, we propose an \textit{Adaptive Maximal Marginal Relevance (A-MMR)} strategy, building upon the traditional MMR~\cite{mmr}.
% We do not use standard MMR because its token selection is highly sensitive to the trade-off hyperparameter, whereas A-MMR avoids such sensitive tuning.
% By evaluating tokens via a unified formulation, A-MMR bypasses the need for hard-coded token splits.
Unlike standard MMR, A-MMR evaluates all tokens under a unified formulation, eliminating the need for hard-coded token splits and sensitive trade-off hyperparameter tuning.
% It dynamically balances the selection process, naturally transitioning from sampling semantically rich, high-attention objects to capturing diverse background representations.
It simultaneously ensures the selection of high-attention semantic entities and their mutual diversity.
Specifically, the selection process is iterative: at each step, we select a token $i$ from the unselected candidate set $\mathcal{U}$ that maximizes the following objective:
\begin{equation}
    i^* = \operatorname*{argmax}_{i \in \mathcal{U}} \left( I_{base}^{(i)} \cdot \left( 1 - \max_{j \in \mathcal{S}} \operatorname{sim}(\mathbf{f}_i, \mathbf{f}_j) \right) \right),
\label{eq:mmr_score}
\end{equation}
where $I_{base}^{(i)}$ represents the attention-based importance score of token $i$, $\mathcal{S}$ is the set of already selected tokens, $\mathbf{f}$ denotes the token feature vector, and $\operatorname{sim}(\cdot, \cdot)$ is the cosine similarity.

The term $(1 - \max_{j \in \mathcal{S}} \operatorname{sim}(\mathbf{f}_i, \mathbf{f}_j))$ represents the \textit{Distinctness} of token $i$. 
This formulation ensures that the algorithm first selects high-attention objects, and then dynamically shifts to cover background context that is semantically distinct from the selected set, which guarantees the diversity of selected vision tokens. 

% \subsection{Phase II: History Frame Pruning (Query-Guided Compression, Stage 3 in Figure~\ref{fig_method})}
\subsection{Token Selection (History)}
For history frames, we filter out redundant information that is irrelevant to the current view. 
We introduce a \textit{Query-Guided Re-weighting} mechanism using the features of the current frame as the query set $\mathcal{Q}$.

First, we calculate the \textit{Spatio-Temporal Relevance} $R^{(i)}$ for each history token $i$:
\begin{equation}
    R^{(i)} = \max_{k \in \mathcal{Q}} \operatorname{sim}(\mathbf{f}_{hist}^{(i)}, \mathbf{f}_{curr}^{(k)}).
\end{equation}
This measures the maximum similarity of a history token to any component of the current view.

Next, we modulate the final importance score to prioritize history tokens that are both originally salient and currently relevant:
\begin{equation}
    I_{final}^{(i)} = I_{base}^{(i)} \cdot \left( \alpha + (1-\alpha) \cdot R^{(i)} \right),
\end{equation}
where $\alpha$ is a balancing factor (we naturally set to 0.5). Finally, we perform the same A-MMR selection (Eq. \ref{eq:mmr_score}) using $I_{final}^{(i)}$ to construct the compact memory pool.

\subsection{Action Prediction}
Upon obtaining this highly informative token set, we feed these through a modality projector and an LLM of the VLA model to predict the sequence of navigation actions. 
Ultimately, this lightweight, plug-and-play pruning formulation elegantly combines detailed visual perception with efficient history compression, fundamentally reducing the computational bottleneck of long-horizon embodied inference.

%% file: tex/4_experiments.tex
\section{Experiment}

\input{fig_tab/tab_main_2}

\subsection{Experimental Setups}

\subsubsection{Datasets}
% For VLN tasks, selecting large-scale, realistic datasets with natural language instructions grounded in visually rich environments is critical. 
In this work we evaluate our method on two public VLN-CE~\cite{vln_ce} benchmarks for evaluating navigation agents: Room-to-Room (R2R)~\cite{r2r}, Room-Across-Room (RxR)~\cite{rxr}. 
% We also evaluate on EnvDrop~\cite{envdrop}.

$\bullet$ \textbf{Room-to-Room (R2R)} was introduced as the well-known benchmark dataset for visually-grounded navigation in indoor environments. 
It is built on top of the \textbf{Matterport3D Simulator}~\cite{matterport3d}, a reinforcement learning environment constructed from high-fidelity panoramic RGB-D scans of real buildings, covering diverse scenes such as homes, offices, and public spaces. 
In R2R, each navigation episode consists of a natural language instruction paired with a corresponding path through a 3D environment; the agent is required to follow the instruction and reach the target location from a given start point. 
% The dataset contains thousands of human-generated navigation instructions with an average length of around 29 words, 
The dataset contains 21,567 open-vocabulary, crowd-sourced navigation instructions, with an average instruction length of 29 words,
and provides a standardized evaluation benchmark for VLN agents to measure success rate and path quality.

$\bullet$ \textbf{Room-Across-Room (RxR)} was developed as a large-scale, multilingual VLN dataset. 
In contrast to R2R, RxR includes over 100K navigation instructions and action demonstrations across multiple languages (e.g., English, Hindi, and Telugu), and each instruction is densely aligned with the agent’s pose traces over time. 
This fine-grained spatio-temporal grounding enhances the role of language in navigation and enables richer supervision for cross-modal learning. 
The diversity in languages and longer, more variable paths make RxR a substantially more challenging and comprehensive benchmark for VLN research.

% $\bullet$ \textbf{EnvDrop} consists of photo-realistic indoor environments represented as viewpoint graphs with panoramic visual observations and human-annotated navigation instructions. 
% At each timestep, the agent receives egocentric panoramic features from its current viewpoint and must select actions conditioned on the instruction and observation history. 
% It provides paired navigation trajectories and natural language instructions, enabling the study of grounded sequential decision-making. 
% It includes diverse indoor scenes with varying layouts and object configurations.

\subsubsection{Baseline Methods}

We compare the proposed approach against the following \textit{training-free} baseline methods, including the latest vision token pruning techniques:

$\bullet$ \textbf{SparseVLM}~\cite{sparsevlm} is a training-free, text-guided token pruning method that estimates visual token importance from the cross-modal self-attention between text and image tokens. 
It then progressively prunes less relevant visual tokens while recycling information into compact representations to improve inference efficiency.

$\bullet$ \textbf{DivPrune}~\cite{divprune} formulates visual token pruning as a ``max-min diversity problem'', aiming to select a subset of visual tokens that maximizes diversity among retained tokens.
In this way, the pruned set covers the original visual information effectively without fine-tuning.
% , leading to robust performance even at high pruning ratios without fine-tuning.

$\bullet$ \textbf{VisPruner}~\cite{vispruner} is a plug-and-play pruning strategy that exploits visual cues rather than relying solely on text–visual attention.
By employing a hard-coded, decoupled selection scheme, it preserves a representative visual subset for model inference, extracting tokens that are both visually informative and semantically distinct.

\subsubsection{Implementation Details}

To ensure a fair comparison, we use identical hyperparameter settings for all baselines and our method.
All experiments are implemented on the StreamVLN~\cite{streamvln}
% ~\footnote{
% The original paper describes a technique termed ``voxel pruning'' to improve inference efficiency. 
% However, at the time of our experiments, neither the implementation details nor the corresponding code were publicly available. 
% We verified that the official codebase does not contain any voxel pruning module. 
% Following standard reproducibility practice, all experiments in this work are conducted strictly using the publicly released official implementation. 
% As a result, voxel pruning is not included in our evaluation, and related details are omitted.
% \Ming{I am not sure if this part is necessary. Spend a lot of words just saying we did not use it. It is not a good practice to say we did not use it. If you really want to explicitly explain this, maybe put into the baseline section, and just say there is no publicly available implementation of voxel pruning.}
% \Qitong{This writing is based on suggestions from Prof. Rasmussen, who suggests that I do some explanation since there is no result or comparison with voxel pruning. I will check with Prof. Rasmussen about the writing and the location of this writing.}
% }
, a recent VLA model specifically designed for vision–language navigation.
Following common practice, we evaluate different pruning levels by setting the pruning ratio to 70\%, 80\%, and 90\% (i.e., pruning the corresponding proportion of visual tokens).
In line with established VLA evaluation settings for VLN~\cite{navila, streamvln}, we perform evaluation on the ``val-unseen'' split of datasets.
Experiments were conducted using Python 3.9, PyTorch 2.8.0, and CUDA 12.8.

\subsection{Validity of Our Method}
\label{sec_4.2}

From Table~\ref{tab_res_main}, we observe that our method consistently outperforms existing pruning methods, and achieves superior performance in most experimental settings. 
Moreover, on both the R2R and RxR datasets, the performance advantage of our method becomes increasingly pronounced as the pruning ratio increases. 
For example, when using SPL as the evaluation metric, our method surpasses SparseVLM, DivPrune, and VisPruner by 5.28\%, 17.81\%, and 7.09\%, respectively, under 90\% pruning ratios on R2R.
We further evidence our superiority over existing vision token pruning methods in the video supplement.

Another notable observation is that under high pruning ratios (e.g., 90\%), several methods, including ours, can achieve a higher Oracle Success Rate (OS) than the unpruned model, and results with relatively low pruning ratio. 
This phenomenon is fully consistent with the definition of VLN evaluation metrics. 
Specifically, OS only requires the agent to enter the goal region at any point along its trajectory, whereas Success Rate (SR) additionally requires the agent to issue the STOP action within the goal threshold~\cite{expl_os1}. 
Therefore, an increase in OS accompanied by a decrease in SR is not contradictory, but instead indicates that the agent successfully reaches the goal region without terminating precisely at the correct location. 
Prior VLN studies, such as~\cite{expl_os2}, have also discussed this gap between OS and SR, showing that a higher OS with lower SR often reflects agents that pass near the goal but fail to stop appropriately. 
Importantly, despite this general phenomenon, our method consistently outperforms existing pruning approaches across most settings, demonstrating its superior ability to preserve task-relevant visual information under aggressive pruning.

\subsection{Ablation Studies}

\input{fig_tab/tab_ablation}

We perform ablation analysis on the R2R dataset, with results summarized in Table~\ref{tab:ablation}.
All the experiments in this section are conducted on the RTX 4090 GPU.

\textbf{Are diversity and semantic importance both necessary for effective vision token pruning in VLN? Yes.}
To better understand the contribution of each component, we conduct ablation studies to evaluate the roles of diversity and semantic importance. 
The results show that both diversity and semantic information are critical for maintaining strong navigation performance. 
When only diversity is considered while ignoring semantic importance, the model tends to preserve visually distinct but task-irrelevant tokens, which weakens its ability to focus on instruction-relevant regions and leads to degraded performance. 
Similarly, when only semantic importance is considered without enforcing diversity, the selected tokens become highly redundant, limiting the model’s ability to capture complementary visual cues necessary for robust navigation.
% As a result, neither diversity-only nor semantics-only pruning yields performance improvements over the baseline, and in some cases even leads to noticeable performance degradation. 
As a result, neither diversity-only nor semantics-only pruning delivers consistent improvements over the original setup. 
In the majority of settings, they result in performance degradation. 
For example, with the pruning ratio of 70\%, the semantic-importance-only setup reduces SR from 52.96\% to 49.43\% and SPL from 48.70\% to 44.83\%. 
While slight gains may appear in isolated cases under specific metrics (e.g., SR, SPL), such improvements are marginal (typically $<$ 1\%) and practically insignificant.
In contrast, our full method, which jointly considers both diversity and semantic importance, consistently achieves the best performance across all evaluation metrics. 
These findings highlight that diversity and semantic importance play complementary roles, and their joint modeling is essential for effective vision token pruning in VLN.

\input{fig_tab/tab_efficiency}

\textbf{Is token merging necessary for VLN? No.}
In the standard vision tasks, some prior token pruning techniques, such as~\cite{sparsevlm}, avoid directly discarding unselected tokens. 
Instead, they merge the pruned tokens with the remaining ones to preserve maximum visual information.
This raises a critical question: 
Is token merging an appropriate and effective strategy for token pruning in the specific context of Vision-Language Navigation (VLN)? 
To answer this, we implemented a variant of our method that merges rather than drops the redundant tokens. 
Interestingly, our experimental results reveal that token merging fails to yield clear performance gains. 
Despite its success in general vision-language modeling, it yields only marginal gains (typically $<1\%$) in isolated cases and frequently degrades navigation performance across most settings.
For instance, with the pruning ratio of 90\%, it reduces SR from 48.40\% to 46.87\% and SPL from 36.51\% to 34.43\%. 
We hypothesize that in VLN scenarios, merging distinct spatio-temporal features can blur fine-grained visual landmarks or introduce noisy context, which ultimately confuses the model's ability to precisely ground directional instructions. 
Therefore, directly discarding redundant tokens proves to be a cleaner and more effective strategy for navigation tasks.

\begin{figure*}[!t]
\vspace*{0.075in}
    \begin{center}
    \includegraphics[width=0.19\textwidth]{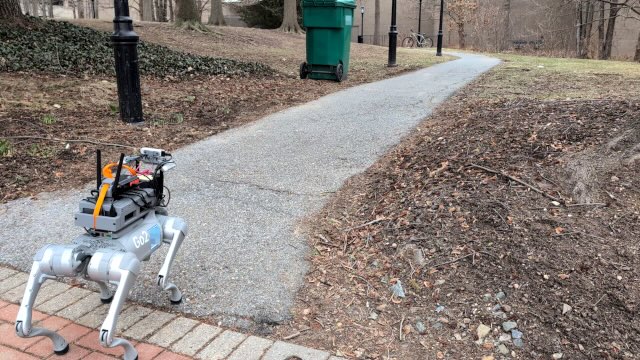}
    \includegraphics[width=0.195\textwidth]{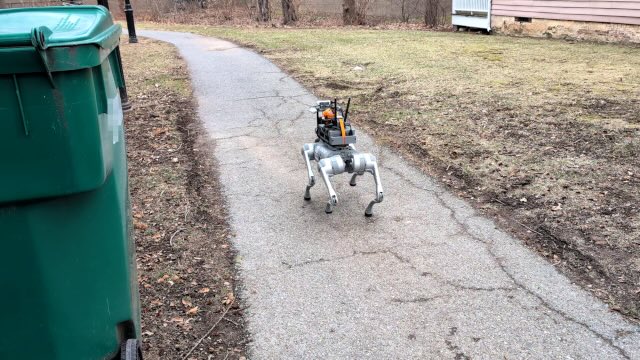}
    \includegraphics[width=0.195\textwidth]{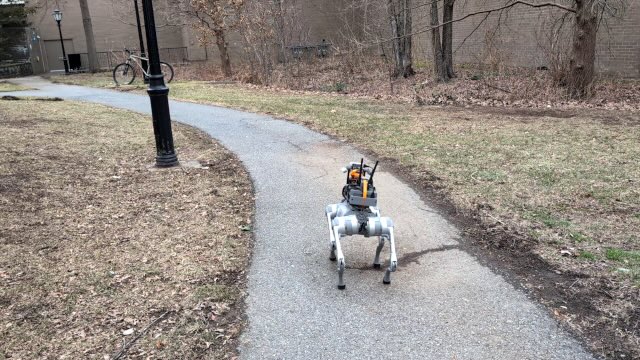}
    \includegraphics[width=0.195\textwidth]{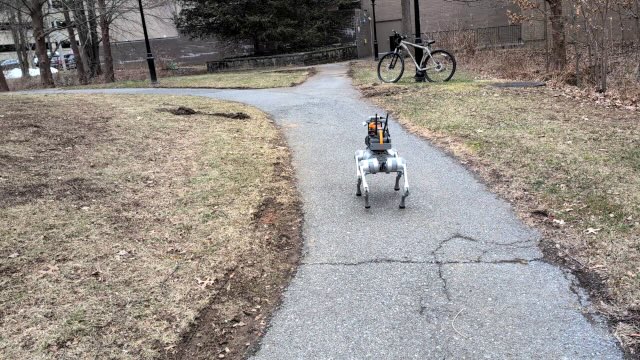}
    \includegraphics[width=0.195\textwidth]{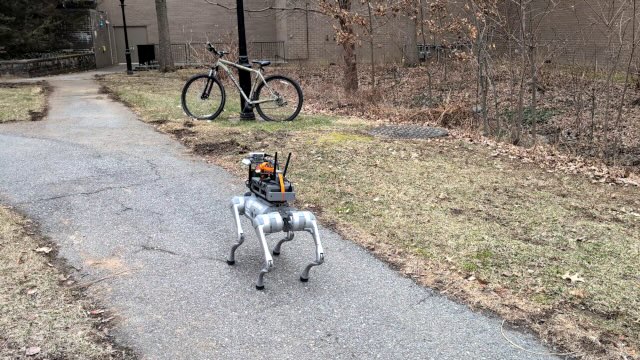}\\
    \end{center}
    \vspace{-5pt} % Add vertical space
    {\footnotesize \faComment[regular]\,\,\texttt{Follow the \textcolor{red}{asphalt path} in front of you past a large \textcolor{red}{trash container}.  Continue along the path until you reach a \textcolor{red}{bicycle} leaning against a black pole.}}
    \begin{center}
    \includegraphics[width=0.19\textwidth]{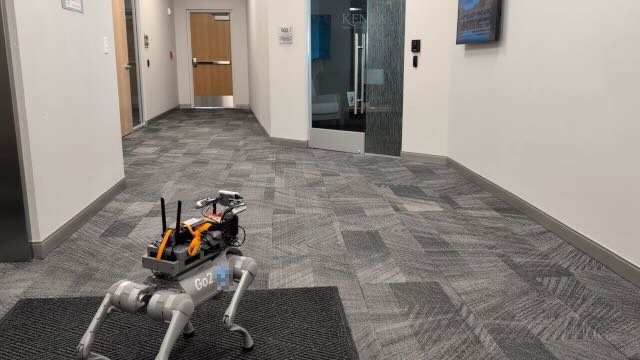}
    \includegraphics[width=0.195\textwidth]{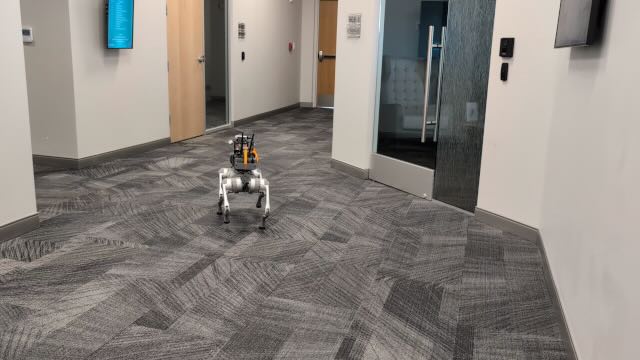}
    \includegraphics[width=0.195\textwidth]{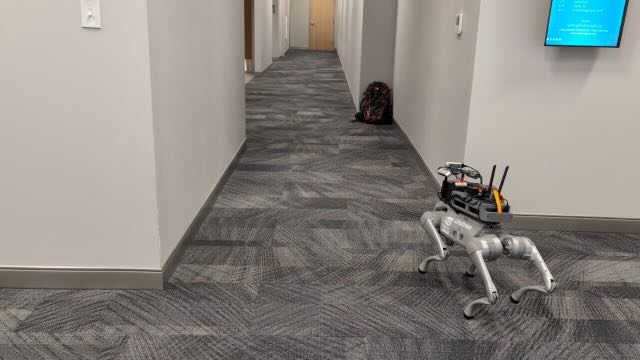}
    \includegraphics[width=0.195\textwidth]{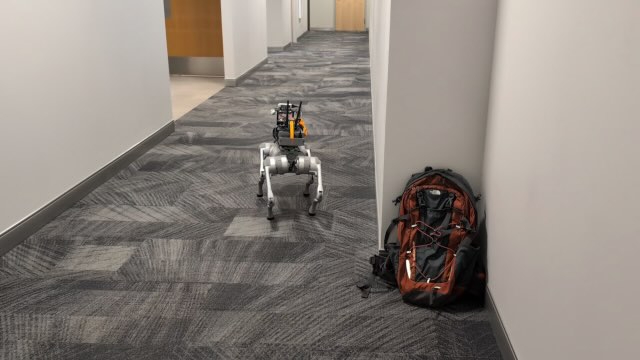}
    \includegraphics[width=0.195\textwidth]{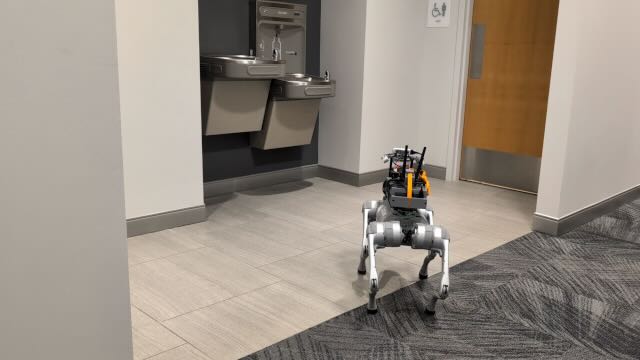}\\
    \end{center}
    \vspace{-5pt} % Add vertical space
    {\footnotesize \faComment[regular]\,\,\texttt{Go to the \textcolor{red}{television} on the wall and turn left.  Go down the hallway on the left and pass a \textcolor{red}{red backpack}.  When you get to a \textcolor{red}{large doorway} on the left, turn into it and go to the \textcolor{red}{drinking fountain}.}}
    \begin{center}
    \includegraphics[width=0.195\textwidth]{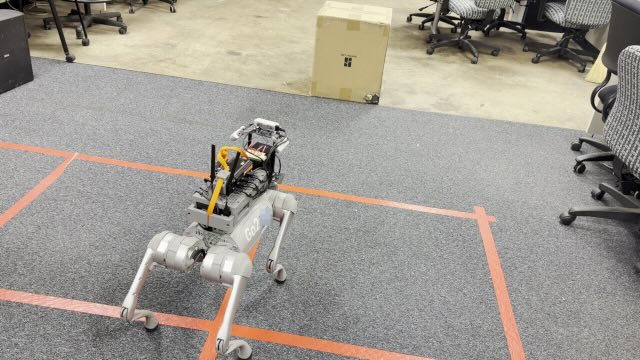}
    \includegraphics[width=0.195\textwidth]{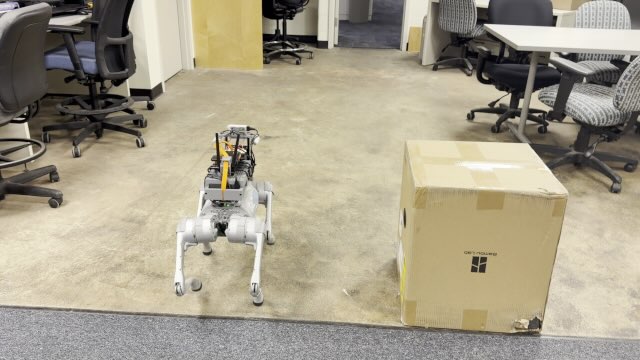}
    \includegraphics[width=0.195\textwidth]{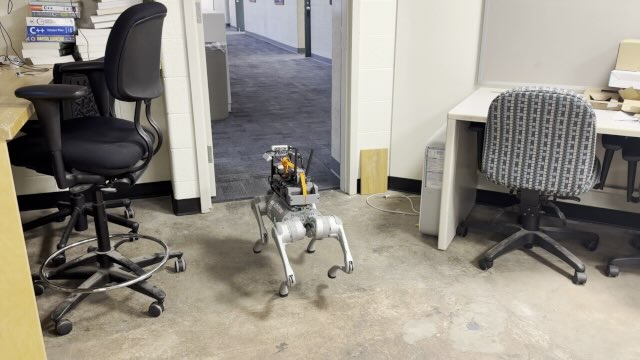}
    \includegraphics[width=0.195\textwidth]{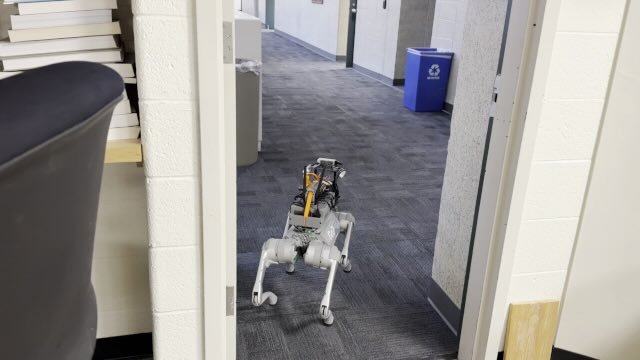}
    \includegraphics[width=0.195\textwidth]{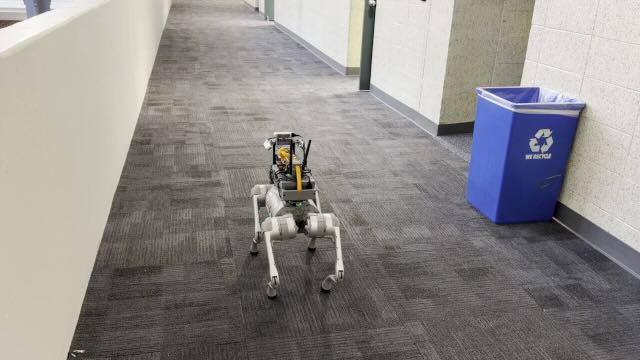}\\
    \end{center}
    \vspace{-5pt} % Add vertical space
    {\footnotesize \faComment[regular]\,\,\texttt{Go around the \textcolor{red}{cardboard box} on the left, and continue straight to the open \textcolor{red}{doorway} ahead.  Go through it completely before heading toward the \textcolor{red}{blue trash container}.  Stop in front of it.}}
    \caption{
    Sample qualitative results demonstrating successful deployment of the StreamVLN model with and without token pruning on a Unitree Go2 quadruped across \texttt{Outdoor}, \texttt{Workspace}, and \texttt{Lab} environments.  
    Note that the landmarks are marked in \textcolor{red}{red}, and the last frame of each sequence represents the VLN-commanded stop location.
    }
    \label{fig_real_world}
\end{figure*}

\subsection{Efficiency Analysis}

Another important aspect of evaluating vision token pruning is to assess its efficiency in addition to performance. 
Therefore, we conduct an efficiency analysis of existing methods on the R2R dataset using an RTX 4090 GPU. 
We adopt a pruning ratio of 90\%, corresponding to an extremely constrained computational budget, to provide a clear comparison of efficiency across different pruning strategies.

From Table~\ref{tab:efficiency}, we first observe that all pruning methods improve inference efficiency compared to the unpruned model under all three efficiency metrics. 
This confirms that vision token pruning is effective in reducing computational overhead. 
Among these methods, our approach achieves the best performance in terms of FPS and latency, or more precisely, performs on par with existing pruning methods while slightly outperforming them. 
This demonstrates that our method offers competitive or superior efficiency under these metrics, while also maintaining stronger task performance as shown in Section~\ref{sec_4.2}.

Another noteworthy observation is that DivPrune achieves significantly lower TFLOPs than all other methods. 
However, as shown in Section~\ref{sec_4.2}, DivPrune yields the weakest navigation performance among the four methods in most cases, particularly when evaluated using SR and SPL. 
For instance, when using SR as the evaluation metric in the R2R dataset, DivPrune underperforms SparseVLM, VisPruner, and our approach by 7.34\%, 13.59\%, and 20.06\%, respectively, under 90\% pruning ratios.
In contrast, our method maintains TFLOPs comparable to other strong baselines while achieving better task performance. 
This highlights that our method achieves a favorable trade-off between efficiency and effectiveness, especially from the FLOPs perspective.

\subsection{Real-world Deployment}

%\textbf{Evaluation Setup} 
\subsubsection{Setup}
We conducted a number of real-world experiments to demonstrate and validate our methods.  
Following \cite{streamvln}, we used a Unitree Go2 quadruped equipped with a forward-facing RGB-D navigation camera (pictured in Figure~\ref{fig_real_world}), but with two key differences.  
First, we ran StreamVLN completely \underline{onboard} the robot on an NVIDIA Jetson Thor T5000, which offers substantial compute and memory resources for an edge platform.  
This allows the robot to operate in the field without the latency of communicating with a remote server (e.g. \cite{streamvln}'s RTX 4090 desktop machine), but most importantly, to carry out instructions when there is no cloud connectivity because of distance, non-line-of-sight, interference, or jamming.  
Second, instead of relying on the Go2's proprioceptive odometry, we obtained more robust visual odometry estimates from 3 hardware-synced cameras via the Isaac ROS Visual SLAM package \cite{isaac-ros-visual-slam}. 

\subsubsection{Results}
The average inference time for a batch of 4 actions was $\sim1.43$ s (0.70 fps) \textit{without} pruning and $\sim1.25$ s (0.80 fps) \textit{with} pruning, but with separate inference and motor controller threads, there were only small pauses in otherwise continuous motion.  
These results are without flash-attention~\cite{flashattention}, which should further speed inference.  
We tried a variety of short to medium-length prompts in several visual environments, both without pruning and with our pruning method using ratios ranging from 50\% to 90\%.  
There was some sensitivity to lighting conditions and prompt wording, but we observed generally good task completion rates.  
The publicly-released version of StreamVLN that we used does not include depth images as model inputs, which we believe contributed to a nontrivial variance in the robot's stopping distance to terminal landmarks.  
Sample results are shown in Fig.~\ref{fig_real_world}.
Please refer to the video supplement for more detailed documentation of the onboard deployment.

%We conduct real-world deployment experiments using a Unitree Go2 robotic dog, equipped with a front-facing camera for live RGB observations and a custom lighter battery to extend operational time. 
%Unlike previous approaches that rely on remote high-end workstations, we deploy our model fully onboard using an NVIDIA Thor edge computing platform (aarch64 architecture). 
%We integrate our proposed vision token pruning mechanism into the StreamVLN codebase. 
%The model outputs four discrete action waypoints per inference step, which are subsequently translated into continuous translation and rotation commands via an integrated client-side PID controller. 
%By utilizing token pruning, we eliminate network communication latency entirely and reduce the average onboard inference time to XXs, enabling smooth, untethered physical deployment.

%\textbf{Real-World Qualitative Results} 
%We evaluate our untethered framework across diverse real-world settings using object-driven navigation prompts (e.g., ``go toward the chair," ``stop in front of the person"). 
%As shown in Figure~\ref{fig_real_world}, the robot successfully grounds these textual prompts to its dynamic visual observations in real time. 
%Our vision token pruning effectively maintains high navigational accuracy while drastically cutting down processing overhead. 
%The system demonstrates robust generalization to novel environments and dynamic obstacles, confirming that our pruning strategy successfully bridges the gap between heavy vision-language models and edge robotics. 

%% file: fig_tab/tab_main_2.tex
\begin{table*}[t]
% \vspace*{0.05in}
\centering

\caption{
Performance comparison of pruning methods on VLN benchmarks.
SR, SPL, OS, NE, and nDTW denote Success Rate, Success weighted by Path Length, Oracle Success Rate, Navigation Error, and normalized Dynamic Time Warping, respectively. 
For each pruning ratio, the best score among all methods is shown in \textbf{bold}.
R2R experiments are conducted on the H100 GPU, while RxR experiments are conducted on the A100 GPU.
$^\dagger$ Although the original StreamVLN paper describes a ``voxel pruning'' technique, at the time of our experiments, its implementation details and code were not publicly available. 
Consequently, our results in this paper strictly exclude voxel pruning to ensure fair comparison.
}
\label{tab_res_main}

\setlength{\tabcolsep}{8pt} 
\renewcommand{\arraystretch}{1.0}

{\fontsize{9.5}{10.2}\selectfont
% \begin{threeparttable}
\begin{tabular}{l|cccc|cccc}
\toprule

\multirow{2}{*}{Methods}

& \multicolumn{4}{c|}{R2R (val-unseen)}
& \multicolumn{4}{c}{RxR (val-unseen)}

\\

& SR (\%) ↑ & SPL (\%) ↑ & OS (\%) ↑ & NE ↓
& SR (\%) ↑ & SPL (\%) ↑ & nDTW (\%) ↑ & NE ↓

\\
\midrule

\rowcolor{gray!50}
\multicolumn{9}{c}{\textit{Retain 729 Tokens (100\%)}} \\

\rowcolor{gray!20}
\midrule

Unpruned$^\dagger$

& 55.74 & 49.66 & 63.60 & 5.06
& 56.53 & 47.26 & 64.14 & 5.71

\\
\midrule

\rowcolor{gray!50}
\multicolumn{9}{c}{\textit{Retain 218 Tokens (↓ 70\%)}} \\
\midrule

SparseVLM
& 46.82 & 42.18 & \textbf{61.83} & \textbf{5.48}
& 40.87 & 35.02 & 53.01 & 7.18
 \\

DivPrune
& 48.78 & 38.14 & 61.34 & 5.52
& 44.48 & 32.15 & 50.62 & 7.29
 \\

VisPruner
& 48.61 & 44.43 & 54.05 & 5.76
& 51.34 & 43.32 & 60.55 & 6.43
 \\

Ours
& \textbf{51.50} & \textbf{47.28} & 58.02 & 5.56
& \textbf{53.19} & \textbf{45.13} & \textbf{61.76} & \textbf{6.26}
 \\

\midrule
\rowcolor{gray!50}
\multicolumn{9}{c}{\textit{Retain 146 Tokens (↓ 80\%)}} \\
\midrule

SparseVLM
& 45.19 & 40.63 & \textbf{60.09} & 5.55
& 34.59 & 30.10 & 52.89 & 7.64
 \\

DivPrune
& 39.31 & 28.64 & 55.14 & 6.47
& 35.13 & 23.92 & 42.68 & 8.49
 \\

VisPruner
& 50.52 & 43.92 & 58.51 & 5.59
& 49.16 & 40.15 & 57.64 & 6.73
 \\

Ours
& \textbf{52.58} & \textbf{46.44} & \textbf{60.09} & \textbf{5.38}
& \textbf{53.36} & \textbf{43.64} & \textbf{61.02} & \textbf{6.16}
 \\

\midrule
\rowcolor{gray!50}
\multicolumn{9}{c}{\textit{Retain 72 Tokens (↓ 90\%)}} \\
\midrule

SparseVLM
& 34.91 & 31.08 & 45.95 & 5.98
& 23.17 & 20.87 & \textbf{50.06} & 8.51
 \\

DivPrune
& 27.57 & 18.55 & 46.06 & 7.36
& 23.22 & 14.56 & 33.73 & 9.83
 \\

VisPruner
& 41.16 & 29.27 & 66.83 & 6.43
& 37.34 & 25.34 & 41.05 & 7.60
 \\

Ours
& \textbf{47.63} & \textbf{36.36} & \textbf{68.46} & \textbf{5.69}
& \textbf{45.71} & \textbf{32.91} & 47.69 & \textbf{6.90}
 \\

\bottomrule
\end{tabular}
% \end{threeparttable}
}

\end{table*}

%% file: fig_tab/tab_ablation.tex
\begin{table}[t]
% \vspace*{0.05in}
\centering

\caption{
Ablation studies of our proposed method.
Note that the original setup of our method is highlighted in \highlightg{light gray background}. 
Similar to Table~\ref{tab_res_main}, SR, SPL, and OS are reported as percentages (\%).
Due to differences in the GPU configuration, the results of the original setup in this table might slightly differ from those reported in Table~\ref{tab_res_main}.
}

\label{tab:ablation}

\setlength{\tabcolsep}{5pt}
\renewcommand{\arraystretch}{1.0}

\begin{tabular}{ccc|cccc}
\toprule

\multicolumn{3}{c|}{\textbf{Setups}} 
& \multicolumn{4}{c}{\textbf{Evaluation Scores}} \\

\cmidrule(lr){1-3}
\cmidrule(lr){4-7}

\textbf{Diversity} & \textbf{Semantics} & \textbf{Merging}
& \textbf{SR} $\uparrow$ & \textbf{SPL} $\uparrow$ & \textbf{OS} $\uparrow$ & \textbf{NE} $\downarrow$ \\

\midrule

\rowcolor{gray!50}
\multicolumn{7}{c}{\textit{Retain 218 Tokens (↓ 70\%)}} \\

\midrule
\rowcolor{gray!20}
\checkmark & \checkmark &  & 52.96 & 48.70 & 58.72 & 5.31 \\

\checkmark & &  & 51.44 & 47.29 & 56.88 & 5.38 \\

 & \checkmark &  & 49.43 & 44.83 & 55.19 & 5.66 \\

\checkmark & \checkmark & \checkmark & 52.69 & 48.37 & 58.13 & 5.34 \\

\midrule

\rowcolor{gray!50}
\multicolumn{7}{c}{\textit{Retain 146 Tokens (↓ 80\%)}} \\

\midrule
\rowcolor{gray!20}
\checkmark & \checkmark &  & 53.29 & 46.89 & 61.56 & 5.31 \\

\checkmark & &  & 54.00 & 47.08 & 60.58 & 5.17 \\

 & \checkmark &  & 48.40 & 42.45 & 57.48 & 5.74 \\

\checkmark & \checkmark & \checkmark & 54.21 & 46.90 & 62.64 & 5.03 \\

\midrule

\rowcolor{gray!50}
\multicolumn{7}{c}{\textit{Retain 72 Tokens (↓ 90\%)}} \\

\midrule
\rowcolor{gray!20}
\checkmark & \checkmark &  & 48.40 & 36.51 & 70.36 & 5.77 \\

\checkmark & &  & 48.02 & 36.18 & 68.95 & 5.75 \\

 & \checkmark &  & 39.53 & 27.80 & 65.14 & 6.58 \\

\checkmark & \checkmark & \checkmark & 46.87 & 34.43 & 70.36 & 5.70 \\

\bottomrule
\end{tabular}

\end{table}

%% file: fig_tab/tab_efficiency.tex
\begin{table}[t]
% \vspace*{0.05in}
\centering
\caption{
Efficiency analysis of inference throughput (FPS), computational complexity (Tera Floating-point Operations Per Second, abbreviated as TFLOPs), and CUDA inference latency (abbreviated as latency) for all the methods.
The best score among all pruning methods is shown in \textbf{bold}.
}
\label{tab:efficiency}

\setlength{\tabcolsep}{12pt}
\renewcommand{\arraystretch}{1.0}

\begin{tabular}{l|ccc}
\toprule
\textbf{Methods} & \textbf{FPS} $\uparrow$ & \textbf{TFLOPs} $\downarrow$ & \textbf{Latency (ms)} $\downarrow$ \\
\midrule

\rowcolor{gray!20}
Unpruned & 4.32 & 10.94 & 231.34 \\

\midrule
SparseVLM & 4.55 & 5.61 & 219.49 \\
DivPrune  & 4.53 & \textbf{1.42} & 220.71 \\
VisPruner & 4.46 & 9.86 & 224.36 \\
Ours      & \textbf{4.68} & 5.61 & \textbf{213.40} \\

\bottomrule
\end{tabular}

\end{table}

%% file: tex/5_conclusion.tex
\section{Conclusion}

In this paper, we addressed the critical latency bottleneck that hinders the real-time deployment of Vision-Language-Action (VLA) models in Vision-Language Navigation (VLN), and introduced a novel, training-free spatio-temporal vision token pruning framework
% Recognizing the unique, history-conditioned nature of VLN tasks, w
% We introduced a novel, training-free spatio-temporal vision token pruning framework. 
Through a unified pruning pipeline that aligns spatial token selection with spatio-temporal memory compression, our method significantly reduces computational redundancy without sacrificing essential navigation context. 
Specifically, we employ an Adaptive Maximal Marginal Relevance (A-MMR) strategy to extract spatial saliency, which then naturally guides a query-based reweighting mechanism for spatio-temporal memory. 
This integrated approach acts as a plug-and-play module that preserves the integrity of pretrained representations.
Extensive evaluations on standard VLN benchmarks demonstrated that our framework predominantly outperforms existing vision token pruning techniques in both navigation accuracy and computational efficiency.
Furthermore, successful real-world deployment on a physical Unitree Go2 quadruped robot validated the robustness and low-latency responsiveness of our approach under practical embodied constraints. 
We hope this work serves as a meaningful step toward bridging the gap between heavy multimodal foundation models and agile, real-time embodied agents. 
% Future work will explore extending this spatio-temporal pruning strategy to multi-camera setups and more dynamic, highly unconstrained environments.